%% file: main.tex
\documentclass[runningheads]{llncs}
\usepackage[T1]{fontenc}
\usepackage{graphicx}
\usepackage{xcolor}
\usepackage{xurl}
\usepackage{float}
\usepackage{tabularx}
\usepackage{hyperref}

\usepackage{float}
\floatstyle{plaintop}
\restylefloat{table}

\DeclareUnicodeCharacter{2212}{-}
\begin{document}
\title{LLM Based Multi-Agent Generation of Semi-structured Documents from Semantic Templates in the Public Administration Domain}
\titlerunning{LLMs Based Generation of Documents from Semantically Similar Examples}
\author{Emanuele Musumeci\orcidID{0009-0004-2359-5032}\inst{1} \and
Michele Brienza\inst{1}\orcidID{0009-0000-1549-9500} \and \\
Vincenzo Suriani\inst{1}\orcidID{0000−0003−1199−8358} \and
Daniele Nardi\inst{1}\orcidID{0000-0001-6606-200X} \and
Domenico Daniele Bloisi\inst{2}\orcidID{0000−0003−0339−8651}}
\authorrunning{E. Musumeci et al.}
\institute{Dept. of Computer, Control, and Management Engineering\\ Sapienza University of Rome, Rome (Italy),
    \email{\{lastname\}@diag.uniroma1.it} \and
UNINT Univeristy, Via Cristoforo Colombo, 200 - 00147 Rome (Italy),
\email{domenico.bloisi@unint.eu}
}
\maketitle       % typeset the header of the contribution
\begin{abstract}
In the last years' digitalization process, the creation and management of documents in various domains, particularly in Public Administration (PA), have become increasingly complex and diverse. This complexity arises from the need to handle a wide range of document types, often characterized by semi-structured forms.
Semi-structured documents present a fixed set of data without a fixed format. %For instance, in the same type of document, the same information may appear in different places or different terms may be used for the same information.
As a consequence, a template-based solution cannot be used, as understanding a document requires the extraction of the data structure. The recent introduction of Large Language Models (LLMs) has enabled the creation of customized text output satisfying user requests.
%Recent iterations of LLMs proved to generate better outputs in few-shot demonstrations, where one or more examples of the desired output are provided to obtain the desired result in contextual specifications regarding the required content and style.
In this work, we propose a novel approach that combines the LLMs with prompt engineering and multi-agent systems for generating new documents
compliant with a desired structure.%, as depicted in Fig. \ref{fig:intro} where the three agents are shown.
The main contribution of this work concerns replacing the commonly used manual prompting with a task description generated by semantic retrieval from an LLM. The potential of this approach is demonstrated through a series of experiments and case studies, showcasing its effectiveness in real-world PA scenarios.
    
\end{abstract}
\section{Introduction}

\begin{figure}[t]
    \centering
    \hspace{-0.5cm}
    \includegraphics[width=0.95\textwidth]{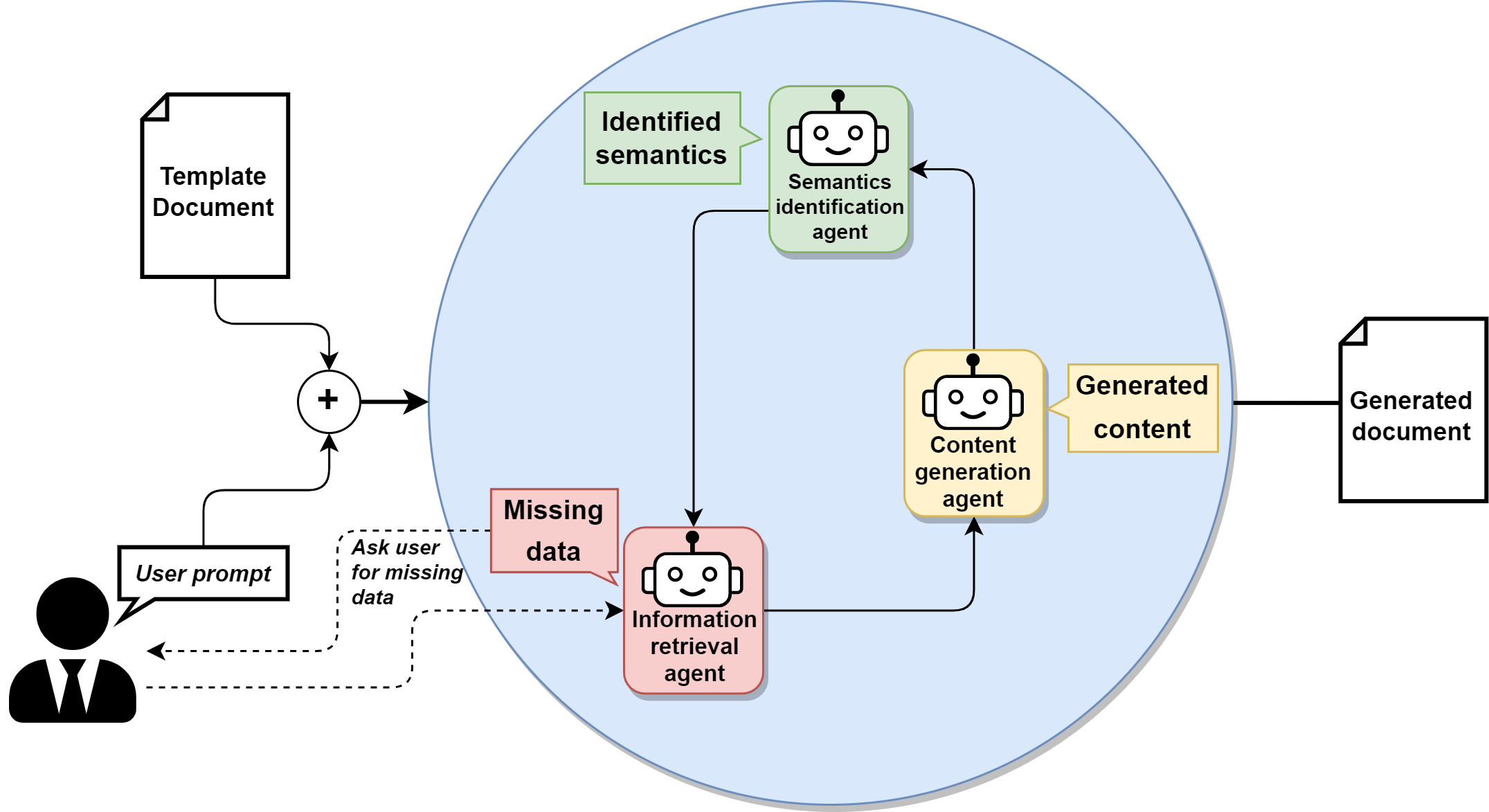}
    \caption{The presented multi-agent architecture with the LLMs used in prompt engineering and multi-agent fashion for generating new documents.}
    \label{fig:intro}
\end{figure}

Document creation is a typical task in the Public Administration (PA) setting, requiring repetitive sub-tasks that offer the potential for automation. For instance, when writing official certificates required in public offices, the required personal information from the requesting user is often very schematic and constitutes only a small percentage of the text present in the document.

The use of pre-made document templates manages only marginally to reduce the effort spent and applies only to rigidly structured documents, where the semantics of missing information can be perfectly defined in the templates themselves. Automation of the writing of this kind of document, which amounts to field-filling, is straightforward. 

Instead, semi-structured documents offer a flexible format, in which missing information cannot be clearly tagged and is usually determined by the semantics of the surrounding context. For instance, standard fields like the current date can appear in different positions in the document with no pre-defined criteria.

Producing this kind of document requires additional effort to adapt their flexible structure to the current use case and the available and required information, with additional effort to recover missing information. The semantics associated with the necessary information require some contextual knowledge, which can be gained by the surrounding context and sometimes the overall semantics of the whole document. For instance, different terms can be used for the qualification of the same field: e.g. "Invoice," "Invoice No." "Bill" and "Purchase Order" can all label the same information (the invoice number).
For this reason, structure and semantics in a semi-structured document are usually intertwined, and developing an efficient document generation pipeline that can run for a wide variety of documents is a very challenging task.

On these premises, the specific task of adapting semi-structured document templates can be supported by the use of Artificial Intelligence and in particular Language Models, to reduce the time spent automating the process and as a means to improve the generality of the automatic approach. 

A straightforward solution could be to generate and therefore refine a document in several incremental steps, with a separate prompt for each one, but this could hinder the quality of the result due to the potential dependencies between data in different parts of the same document, that might be located in structurally unrelated sections. A fully unsupervised approach would therefore run the risk of incurring in hallucination due to the limited effectiveness of long context windows for long document structures. 

Prompt engineering guidelines usually require providing preliminary and accurate context to the role of the LLM agent in the required task. Therefore it is possible to alter and improve the result on the same user request by prepending an accurate description of the role of the LLM agent to the actual user prompt. 

Under these assumptions, it is possible to improve the incremental trial-and-error document generation by introducing several agents, each with a defined fine-grained role in the generation process. In this framework, the capabilities of each agent can be tuned by providing it with an accurate description of its role. Moreover, each agent can be augmented with a memory component local to the agent, sampled and applied specifically for its role, and with additional capabilities that compensate for the lacks of Language Models, such as interaction with the World Wide Web in real-time, access to private custom knowledge bases and information feeds to enhance agents with domain-specific knowledge, or the execution of specialized code as in a Function-as-a-Service framework. This kind of architecture is perfectly compatible with the emergent AI-as-a-Service (AIaaS) paradigm\cite{cobbe2021artificial}.

The multi-agent process assists the user by iteratively refining the prompt with the guidance of a pre-existing structure extracted from similar documents. Then, context-specific prompts are provided to the various agents during the process, depending on their specific role and the original document structure, with little to no human supervision. Interaction between the agents, which is conversational in nature, can include direct interaction with the user in cases in which intervention is needed. %The augmentation of agents and orchestration of such interaction will be a subject of future research.

We present a workflow and interaction framework for the LLM-assisted multi-agent generation of a semi-structured document in the PA domain, with limited human supervision. We then show the prompt refinement process necessary to obtain the required results for each specific agent role in the current workflow. \\
The code and the additional results obtained from this work can be found at the following webpage \url{https://sites.google.com/uniroma1.it/multi-agent-documentgeneration/home-page}.

The remainder of the paper is organized as follows. Section \ref{sec:relatedwork} contains a description of the state of the art. Section \ref{sec:proposedapproach} presents the description of the workflow and interaction framework. The prompt refinement process is then shown in Section \ref{sec:results}. Finally, conclusions are drawn in Section \ref{sec:conclusions}.
\section{Related Work}
\label{sec:relatedwork}

Since their release to the public, Large Language Models (LLMs) have shown great potential for a wide range of daily tasks \cite{radford2019language}. In particular, their capabilities in document editing and generation use-cases offer great potential for their successful application to the PA domain. Most LLMs have shown greater performance in zero-shot \cite{NEURIPS2022_8bb0d291} and in particular few-shot\cite{pmlr-v206-hegselmann23a} tasks, where examples of acceptable results are provided along with the instructions for the task to be executed.

It has been shown that better results can be obtained by applying guidelines for prompt engineering \cite{white2023prompt}. Usually, when a human is involved, the prompt refinement becomes a trial-and-error process, by improving the final result by incremental changes to the initial prompt, based on the generated output. 

Prompt engineering proved to be a crucial step both for the average and advanced users in applications of LLMs to the production of semi-structured documents, where the original structure requires subsequent adaptations through a trial-and-error process. Through prompt engineering, it is possible to improve the quality of the output obtained and especially its compliance with contextual specifications regarding the required content and style. For this reason, the main challenge of allowing PA entities to successfully integrate LLMs in their workflows is to enable the inexperienced user to create efficient prompts and to minimize the time spent improving the task description provided as a prompt to the Large Language Model.

As a trade-off for their versatility, LLMs incur in the problem of hallucination, causing results to be skewed and inaccurate or biased with respect to the original requests, especially with longer context windows. In the document generation task, especially for the generation of longer documents, prompts might tend to be long and rich in information, with the risk of causing hallucinations. 

\subsection{LLMs in the PA Domain}

The integration of Large Language Models (LLMs) in automating document generation processes, particularly in the domain of PA, has seen significant interest due to the amount of document manipulation required in this domain. Some works are helping in information extraction from those documents, for example, when dealing with extracting and classifying relations from tenders of the PA \cite{siciliani2023oie4pa}.

Prior studies are nowadays predominantly focused on leveraging LLMs for structured data extraction, text summarization, and content customization \cite{li2022pretrained} to enhance administrative efficiency and user engagement. Prompt engineering has shown relevant results in improving the LLM's generation capabilities\cite{white2023prompt}, and, with guiding principles, LLMs can meet requirements and allow for enhanced quality in response \cite{bsharat2024principled}.

An approach for information extraction for unstructured documents is presented in \cite{peng2023embedding}, where an embedding-based retrieval system with LLM is used for effective agriculture information extraction from unstructured data. The system features an embedding-based retrieval system along with LLM question-answering to automatically extract entities and attributes from the documents, and transform them into structured data.

When dealing with novel documents, also Retrieval-augmented generation (RAG) approaches allow large language models (LLM) to retrieve relevant knowledge, showing promising potential in mitigating LLM hallucinations and enhancing response quality, and, chance, facilitating the adoption of LLMs in practice \cite{chen2023benchmarking}. However, existing RAG systems are often inadequate in answering multi-hop queries, which require retrieving and reasoning over iteratively. An improvement of this technology has been proposed in \cite{tang2024multihop} where multi-hop reasoning steps are introduced in the RAG system. %In \cite{wang2023docllm}

The difficulties are even more common when dealing with the application of LLMs in generating semi-structured documents from semantically similar examples. This remains relatively unsolved and they still struggle with tasks that require generating complex, structured outputs \cite{tang2023struc}. This gap is primarily due to the inherent complexity of semi-structured documents, which defy conventional template-based approaches. Several approaches have been proposed to handle them.
A notable example is represented by \cite{NEURIPS2022_9d560961}, where Chain-Of-Thought is presented as a series of intermediate reasoning steps that significantly improve the ability of large language models to perform complex reasoning. In particular, it is shown how such reasoning abilities emerge naturally in sufficiently large language models.

A closer step in the generation of semi-structured documents is represented by the Directional Stimulus Prompting Technology \cite{li2023guiding}, where the LLM output is conditioned to generate desired outcomes, such as including specific keywords.
The research on semantic understanding and context-aware generation provides foundational insights but stops short of addressing the specific challenges posed by semi-structured documents in PA.

Furthermore, the Artificial Intelligence as a Service ('AIaaS') trend \cite{cobbe2021artificial} is going to play a growing role in society's technological infrastructure, enabling, facilitating, and underpinning functionality in many applications. AIaaS providers therefore hold significant power at this infrastructural level and with the upcoming legislations in Europe, their role can easily be diffused in the workflow of the public offices. 
The \textit{AIaaS} approach aligns also with our proposed multi-agent approach in document generation tasks. The multi-agent approach has been demonstrated to improve problem-solving in overcoming the limitations of individual models \cite{rasal2024llm}. 
In the PA domain, this distributed approach offers modularity and scalability, potentially suitable for handling various document types and complexities within PA settings.

In summary, while the literature provides valuable perspectives on the capabilities and applications of LLMs in various contexts, our work contributes a novel methodology and interaction framework for the LLM-assisted semi-structured document generation in PA, including multi-agent assistance in document generation and paving the way for further innovation and exploration in this domain.

\section{Proposed Approach}
\label{sec:proposedapproach} 
The usual interaction model between an LLM and an inexperienced user for document generation features a trial-and-error process, aimed at refining the prompt until a satisfying result is reached. Longer prompts are required for longer documents, making it even more difficult to obtain a satisfying result.
Moreover, in detail-rich documents, the performance of LLM agents is bound to decrease when document generation requires many tasks to be completed.  

Given the requirement for user supervision in a trial-and-error process, we propose a different workflow aimed at minimizing user intervention. The proposed process, iterative in nature, follows the overall structure of a document template, which can be extracted from a pre-existing template document provided by the user as input.

The user is allowed to provide an initial prompt to describe the overall expected result. The initial prompt is then refined throughout iterations, to hold all the missing data required to generate the document, accumulating in the original prompt the outcomes of user interventions whenever requested by the agents. The \emph{accumulated prompt} will serve as a data source throughout the document generation steps.
Following the structure of the template document, the output document is then generated section-by-section, in reading order.

During the generic generation step, LLM agents are interrogated to solve fine-grained tasks depending on the availability of the information required to generate semantically suitable content for the current section, according to the semantics provided in the corresponding section in the template document. Each agent is instructed with a previously engineered prompt, describing its task, which is then completed by context-dependent information, depending on the semantics of the current section and the data extracted from the \emph{accumulated prompt}.

The multi-agent framework allows specializing agents as much as necessary to prevent hallucinations, in sections where available contextual pieces of information are prone to provide undesired results (like what could happen in case the provided context is very short or the text is very schematic). Post-processing may be applied to improve the results, especially if a schematic output is expected, by explicitly asking the LLM agents to return specific tokens in case some conditions are met, as a way to force them to not hallucinate and comply with their role in the workflow. Detecting these tokens in the output may help in managing limit cases that would otherwise disrupt the workflow, improving the overall  system robustness.

User intervention is required only in case the pieces of information retrievable from the \emph{accumulated prompt} are not enough to comply with the semantics of the current document section, so the frequency of user intervention depends on the quality of the initial prompt. The advantage of this approach is that the \emph{accumulated prompt} is used only as a data source: in this way, the agent tasked with extracting data from the prompt is less prone to hallucinate when the user prompt is not complete or clear. After user intervention, the new data provided by the user is added to the \emph{accumulated prompt}, to be stored for future retrievals.

\begin{figure}[t]
    \centering
    \hspace{-0.5cm}
    \includegraphics[width=\textwidth]{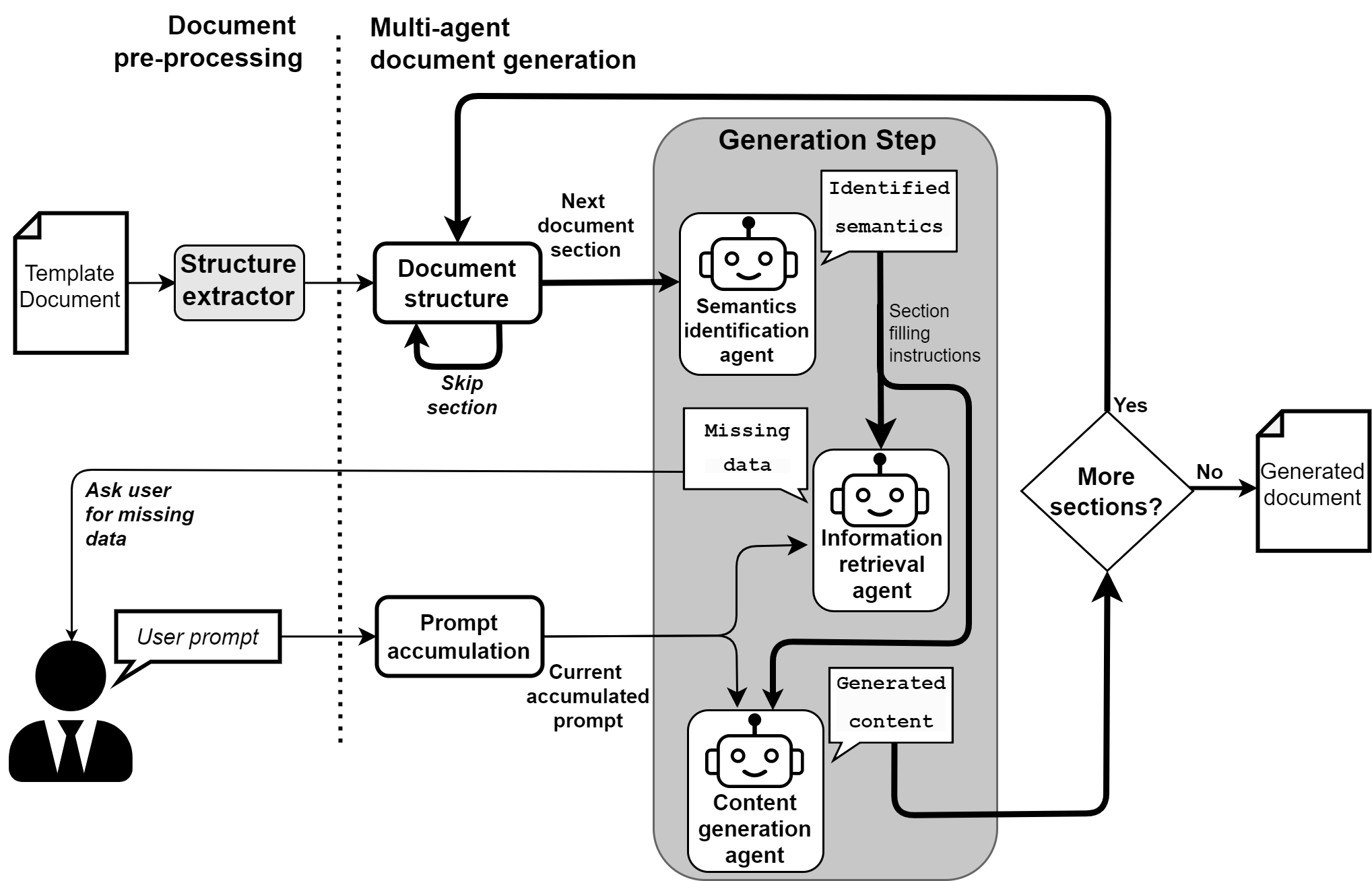}
    \caption{Representation of our multi-agent architecture. The workflow for the generic generation step is highlighted by the bold black arrows.}
    \label{fig:architecture}
\end{figure}

To ensure flexibility in the emulation of the original document template, the user is allowed to optionally skip the generation of a document section at any time, leaving the accumulated prompt unaltered.
A representation of the workflow obtained according to this interaction model is shown in Fig. \ref{fig:architecture}.

Although our work focuses on the components atomically necessary to generate a document section-by-section, the presented interaction model allows the integration of additional agents to manage different aspects of document editing, depending on the level of structuring of the document and the level of expertise required by the specific task assigned to the agent.

\subsection{Template Pre-processing}
\label{sub_sec:template-preprocessing}
The document structure can be extracted on a format-dependent basis, using pre-existing tools. In our case, we used a REST API interface for cloud-based processing using the Adobe Extraction API\footnote{\url{https://developer.adobe.com/document-services/docs/overview/pdf-extract-api/}}, to extract bounding boxes and contents from figures, text blocks, and tables in the document. It is not important to deduce the field semantics at this stage as we only need structural cues for the next steps.

\subsection{Multi-agent Interaction}
The user is invited to provide an initial prompt giving an overall description and directives for the generated document, such as more general qualities like style or tone of the text or more specific information and data necessary for document generation. It is possible to leave the initial prompt empty, in which case the maximum level of user intervention will be required throughout the generation process.

The current workflow features a set of three LLM-based agents, each corresponding to a phase of the generic content generation step for a single document section, in order: \emph{Section Semantics Identification}, \emph{Information Retrieval}, \emph{Content Generation}.

The \emph{Section Semantics Identification} step is aimed at identifying the semantics of the current section from the document template. In case the template section contains tokens that need to be replaced in the current document section, which can appear as placeholders, such as "Name", "Surname", "Birthday", "City" or explicitly already populated with data, such as "John", "Doe", "01/01/1970", "Washington", the usual Natural Language Processing pipeline to perform Entity recognition and Semantic parsing on a sentence would generally require performing several preliminary tasks such as Part-of-Speech tagging, Named Entity Recognition, Relationship Extraction, before the actual semantic tagging of tokens, to build a semantic representation of the sentence good enough to extract the tokens of interest correctly.
Using LLMs for this task allows instead exploiting their Commonsense Knowledge \cite{krause2023commonsense}. This step is therefore managed by the first LLM agent, the \emph{Semantics Identification agent}, which autonomously identifies the semantics of the current section from the template document, identifying replaceable data in the provided template section.

The agent output is a list of directives and instructions on how to reproduce the semantics of the corresponding template, whenever the text allows deducing it, serving as instructions for the \emph{Content Generation} phase, along with a list of identified replaceable data. In case the semantics of any of such data are identified, the list of instructions will contain directives on how to add them to the text generated for the corresponding section in the output document. Using only a schematic representation of the semantics as an output causes a degradation in the quality of the generated content downstream, therefore the instructions provided by the agent are discursive and verbose. The output of this phase will be provided to the \emph{Information Retrieval agent}, to identify data required in the current section, and the \emph{Content Generation agent}, enabling it to reconstruct the semantics of this template section in the output document.

It should be noted that the list of replaceable data might contain data that is already available in the accumulated user prompt as well as missing data. 
For this reason, the second phase, destined to \emph{Information Retrieval}, is aimed at using the available information to retrieve the data specifically required by the current document section, according to the semantic cues previously extracted.

The second agent, the \emph{Information Retrieval agent}, is specialized in extracting the required information from the accumulated prompt, which at the first step coincides with the initial prompt, and determining which data could not be retrieved. In case it is not possible to find all the required data (according to the instructions from the \emph{Semantics Identification agent}), user intervention is required, to specify through a textual prompt the actual replacement values for the missing data. The result of this interaction is added to the \emph{accumulated prompt}, to be used by the \emph{Content Generation agent} as a data source, or to be stored for the \emph{Information Retrieval} phase of later iterations. Therefore, if the user decides not to specify some information, they will not be integrated into the result and will be ignored in the output content.

As a last step, during the \emph{Content Generation } phase, the \emph{Content Generation agent} is therefore instructed to generate the textual content for the current document section, using the \emph{accumulated prompt}, now enriched with the required information, and the instructions coming from the \emph{Semantics Identification agent}.

\begin{figure}[t]
    \centering
    %\hspace{-0.5cm}
    \includegraphics[width=0.95\textwidth]{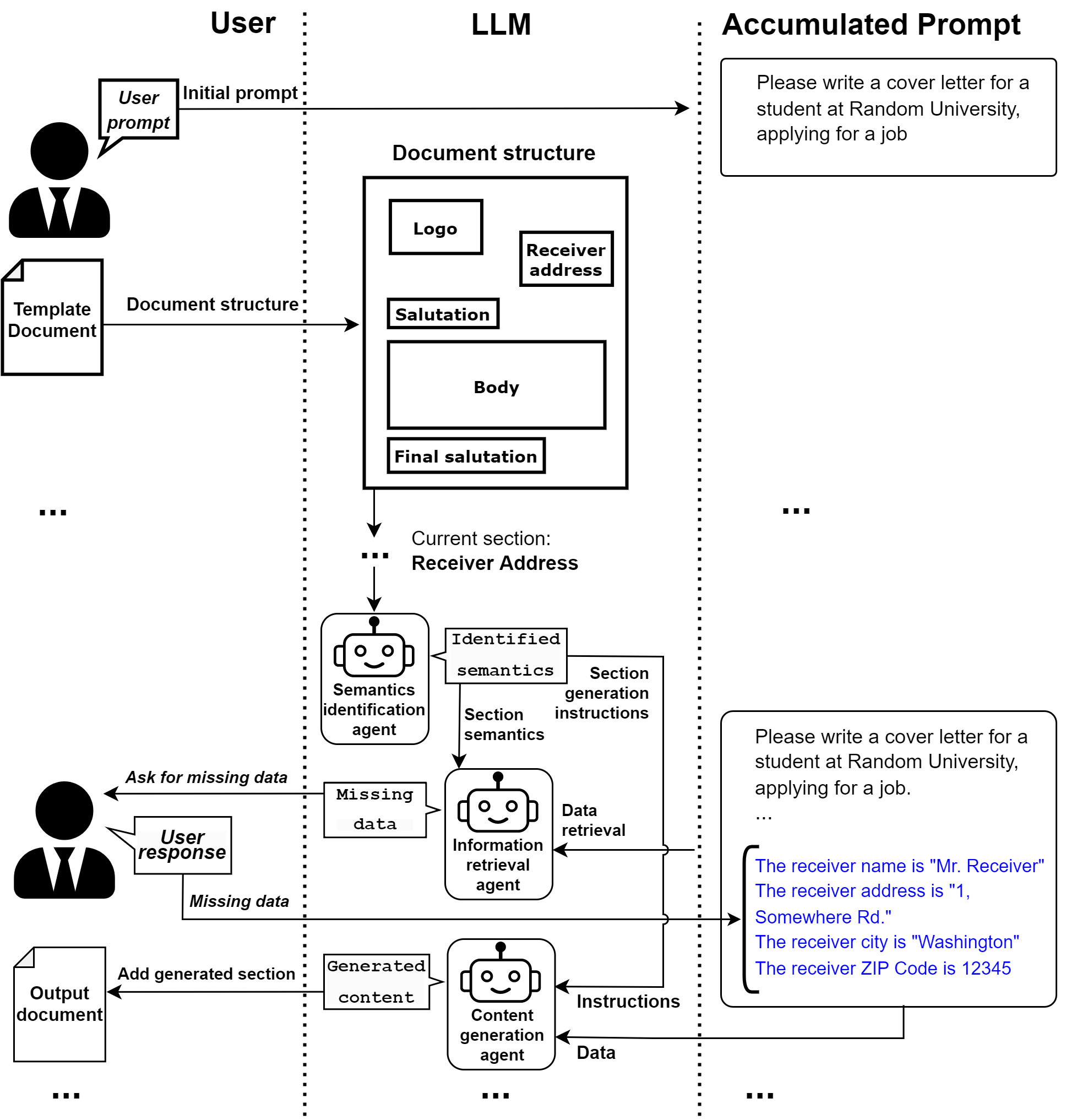}
    \caption{Representation of a generation step instance. Notice how the \emph{accumulated prompt} is enriched with the missing data provided by the user.}
    \label{fig:workflow_instance}
\end{figure}

At the end, both an output document and a refined prompt are obtained. In particular, the refined prompt will contain all the missing data identified throughout the generation process. An example of a workflow instance is shown in Fig. \ref{fig:workflow_instance}.

\subsection{Document Post-processing}
During the document generation, only text content is processed, while the same structure of the original template document is used, including figures and other graphical elements, which can be skipped by the user during the generation process to avoid including them. Thanks to the modularity of this framework, additional agents can be added to improve the graphical appearance and improve its dependence on the context and the semantics of the user requirements, and additional processing steps can be added downstream to improve the graphical appearance of the result (for example by generating context-dependent images) but this is outside of the scope of the current work.

\section{Experimental Evaluation}
\label{sec:results}
Agents were built using the latest OpenAI \emph{GPT 3.5 Turbo} model (\emph{gpt-3.5-turbo-1106}) with a context window of 16,385 tokens and a maximum of 4,096 output tokens.

\input{table_field_1}

Experiments are aimed at designing an effective prompt for each agent, by performing a trial-and-error process to obtain a correct and contextually appropriate response from the LLMs.

The evaluation process tests the agents' ability to understand the assigned tasks, fine-tuning each agent's ability to conform to the expected outputs,  given their specific role in the generation workflow.

The process of crafting an effective prompt that maximizes conformity of the output of a Large Language Model to the original requirements has the potential to dramatically improve the quality of the generated output, especially when this output is used in an intermediate step of a processing pipeline.
The baseline version of the prompt point is obtained by following general prompt engineering guidelines \cite{white2023prompt}, the most important one consisting of writing in a clear and imperative tone the task assignment for the LLM. Starting from the baseline and analyzing the response, it is then possible to refine the prompt in incremental steps, by adding instructions to force the model toward a more desirable output.

The most important undesirable behavior to keep under control is the chance for hallucination, which in this case can alter the data in the generated content and therefore provide false and unsatisfying textual content for the document section being generated.

Each prompt consists of two components: a "\emph{system prompt}" containing the pre-assigned instructions used to inform the agent of its specific task, instructing it with its role and the guidelines for the generation of its output, and a "\emph{context-specific prompt}", which contains the actual context-specific text for the current instance of the agent's task.

Obtaining a robustly engineered prompt requires careful tuning, considering the variety of possible inputs in a semi-structured document. On one hand, very schematic text consisting of single entities (such as "First name", "Last name" and other specific data) poses a challenge and requires more carefully engineered prompts. On the other hand, output should still be desirable in the case of less schematic, more discursive sections.
The generic input for the generation step, which will be forwarded to the \emph{Semantics identification agent}, is the textual content of the currently processed template document section, which is necessary to obtain semantic cues about the desired output text.

The output of the generation step coincides with the output from the \emph{Content generation agent}.

\subsection{Semantics Identification Agent}
The \emph{Semantics Identification agent} is tasked with producing a discursive and fluent explanation of the semantics of the current document section and most importantly of the instructions and directives for the generation of the content to be included in the output document, based on the expected semantics.
To avoid hallucinations it is extremely important to have a non-schematic and unspecific description of the semantics of the current document section.

We show the prompt engineering process based on the answers of the same agent to different \emph{system prompts}, progressively refined in incremental steps, for different template texts, in order of increasing verbosity of the \emph{context-specific prompt} (therefore in decreasing order of schematism).

The first more schematic template text is "\emph{Your name}", as could be found in any template document for a generic letter, in a field where the name of the user is required. The relevant results for this refinement process are provided  in Tables \ref{tab:semantics-identification-table-1}.\\ 
\input{table_field_2}

The initial baseline version of the prompt describes the agent's task without specifying that the output should contain instructions on how to reproduce the semantics of the provided template text.
The agent response is heavily affected by hallucinations, answering with plausible text for a generic letter, without giving a generic description of its semantics, clearly missing its objective. From the analysis of the answer, it can be assumed that the agent misinterprets the request "\emph{Respond with the content that could be written}" as a request to provide the actual content of the generated document.

\input{table_field_3}

\input{table_missing_1}

The addition of a clear instruction "\emph{Give just the action to do}" in the second version of the system prompt definition gives the agent a clear explanation on how to perform its task. This sentence is crucial to improve the conformity of the agent's response to its original task, instructing it to provide a description of the action to perform to reproduce the semantics of the template section.

The relevant results for template text "\emph{Dear Mr./Ms.(Lastname)}" are provided in Table \ref{tab:semantics-identification-table-2}. Here, the output to the second version of the prompt could be considered correct but it could be even more synthetic, considering that the optimal response would just be and instruction to "\emph{Add a salutation}".

\input{table_missing_2}
Given that we aim to maintain the agent output as short and synthetic as possible, we found out that the result can be further improved by providing examples of correct and desirable outputs. This 'few-shot prompting' technique allows constraining the output to adhere as much as possible to a desirable format, furthermore improving the understanding of the agent on how to better perform in the assigned task. As seen in the third version of the prompt in Tables \ref{tab:semantics-identification-table-1}, \ref{tab:semantics-identification-table-2} and \ref{tab:semantics-identification-table-3} an addition of examples in the prompt makes it possible to obtain an answer that fits the desired requirements.

In particular, it should be noted how the tone of the agent's answer in \ref{tab:semantics-identification-table-3} becomes increasingly descriptive and imperative, which is nearer to the required style, as the task prompt is refined.

\input{table_missing_3}

\subsection{Information Retrieval Agent}
The \emph{Information Retrieval Agent} is tasked with retrieving information from the context initially provided by the user and progressively accumulated throughout iterations in the \emph{accumulated prompt}, according to the instructions of the previous agent. The task description for this agent is obtained from the answer of the first agent as seen in Tables \ref{tab:missing_information_1}, \ref{tab:missing_information_2} and \ref{tab:missing_information_3}.

The agent is instructed to return the \verb|[ALL_INFO]| token to signal that it has managed to retrieve all the required information from the existing accumulated prompt.

In Table \ref{tab:missing_information_1}, when asked for the name using the baseline version of the prompt, the agent does not reply with the \verb|[ALL_INFO]| token but requests further information, even though in the current \emph{Accumulated prompt} the name of the user is already available, clearly deviating from the assigned task, while in Table \ref{tab:missing_information_2}, the agent erroneously replies to the baseline prompt with a further request for information, including context information that does not concern its task. The prompt is therefore refined by adding a clear example-based instruction  ("\emph{Your output must be like: ...}"), constraining the model to conform to a specific output format. Moreover, a reinforcement instruction is added to avoid hallucinations ("\emph{Strictly respond with only the information that is missing.}").\\
A particularly interesting study case consists of using the engineered prompt and providing a text without a semantic value, such as a placeholder text:
\\\\
\textbf{Template text:}
\emph{
Lorem ipsum dolor sit amet, consectetur adipiscing elit...}
\\\\
The template text is the placeholder text ("\emph{Lorem Ipsum ...}") typically used to test graphical templates. The agent answer perfectly conforms to the prescribed task as this placeholder is usually found in the discursive sections of document templates:
\\\\
\textbf{Agent Answer}: \emph{The missing information to satisfy the request is the content of the main body of the document.}
\\\\
\subsection{Content Generation Agent}
The \emph{Content Generation Agent} is tasked with generating the text in the output document following the instructions of the other agents and using the information contained in the \emph{accumulated prompt}.\\
The \emph{system prompt} for this agent, as in the cases of the previous agents, contains clear instructions and constraints to obtain a desirable output.
The baseline prompt is: 
\\\\
\textbf{Agent task}:

\emph{
You are an assistant with the purpose of generating a document
with the available information.
You have the following information:\newline
}
\verb|{Accumulated prompt}|
\newline
\emph{
For example, if you read "Add salutation name", write the salutation only.
}
\\\\
where \verb|{Accumulated prompt}| is the currently \emph{accumulated prompt}.\\
With this baseline, the model tends to add initial greetings and a conclusion. This deviation from the original task could be solved by forbidding the addition of out-of-context information, by adding the following sentence
\\\\
\emph{
Remember that in an opening paragraph, it is absolutely forbidden
to write "dear someone", or "sincerely" at the end of the document.
}
\\\\
As for the previous agents, adding examples of the desired result drives the agent's output to the desired format.

\begin{figure}[t]
    \centering
    \includegraphics[width=0.93\textwidth]{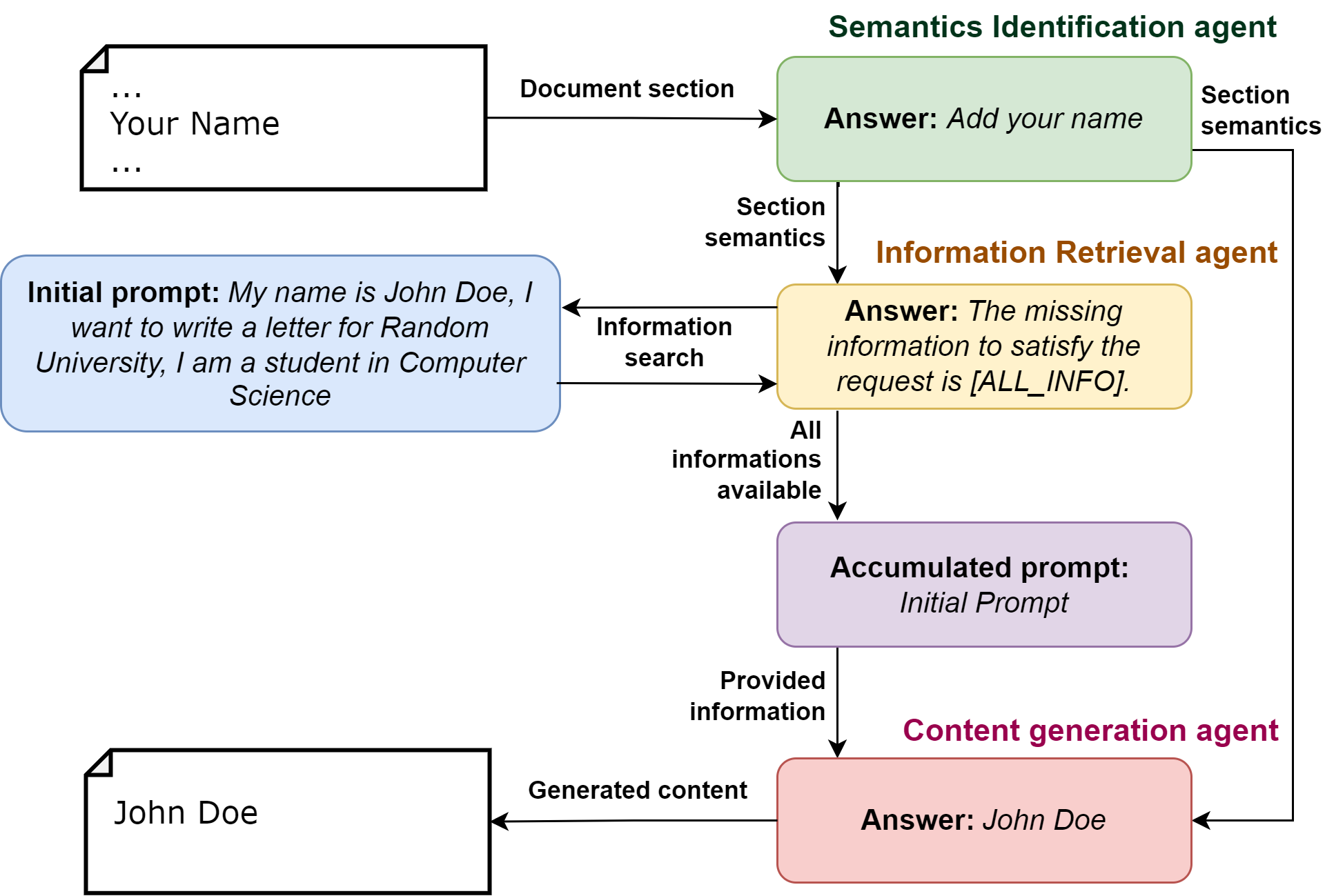}
    \caption{Agent responses throughout a single generation step, starting from the template text: "\emph{Your name}". The figure represents a single generation step for a section of the document, showing an example of successful generation using the engineered prompt. In this case all information is retrieved in the original user prompt, so no information is asked to the user.}
    \label{fig:content_filling_name}
\end{figure}

\subsection{Prompt-engineered results}
\begin{figure}[t]
    %\vspace{10mm}
    \centering
    \includegraphics[width=0.9\textwidth]{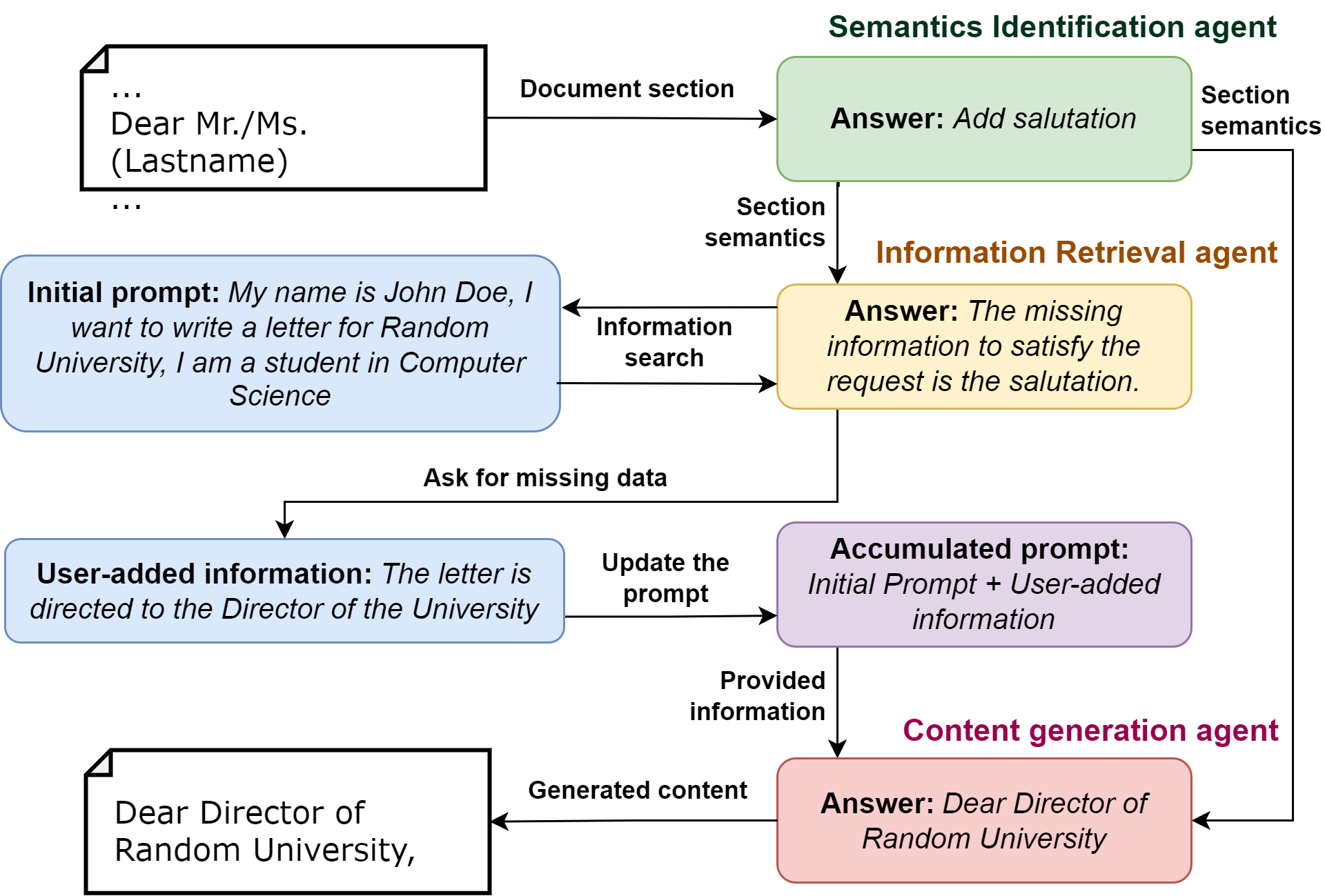}
    \caption{Agent responses throughout a single generation step, starting from the template text: "\emph{Dear Mr./Ms.(Lastname)}". In this case, user intervention is required to add missing information.}
    \label{fig:content_filling_dear}
\end{figure}

Examples of a full processing iteration of template sections with the refined prompts are shown in Figures \ref{fig:content_filling_name} and \ref{fig:content_filling_dear}.

Fig. \ref{fig:content_filling_name} in particular presents a case in which, thanks to the prompt engineering process, the \emph{Content Generation agent} is able to retrieve the necessary data from the \emph{accumulated prompt} without asking for user intervention, then using the instructions for the reproduction of the semantics of the document section, provided by the \emph{Semantics Identification agent}, to generate a desirable result.

Fig. \ref{fig:content_filling_dear} instead presents a case in which some missing information is detected, which requires user intervention to provide such missing data, which is then added to the \emph{accumulated prompt}.

In both cases, a dramatic improvement in the quality of the agents' responses was obtained by adding examples for the structure of the expected result.

%The forbidden rule added to the prompt has made it possible to obtain the result in Figure \ref{fig:content_filling_body}, which contains just the content of the text without any salutation or conclusion, frequently occurring error without having specified the rule. Furthermore, with the support of the field identification agent, the content generated respects the semantic value explained by the description of the template field.

\section{Conclusions}
\label{sec:conclusions}

This work presents a workflow and an interaction framework for the LLM-assisted multi-agent generation of a semi-structured document in the PA domain, with limited human supervision. 

We exploited the capabilities of a Large Language Model, taking advantage of a multi-agent architecture, involving three roles, namely, a \textit{Semantics Identification Agent}, an \textit{Information Retrieval Agent}, and, a \textit{Content Generation Agent}. With the resulting architecture, we highlighted the necessary prompt refinement process to obtain desirable results for each specific agent role in the presented workflow. The approach is successful in both schematic and verbose document sections providing enough cues about their semantics.

Example-based document generation proves to be perfectly adaptable to the PA scenario, where a progressive transition to Artificial Intelligence tools is expected in the next years.

The paradigm shown in the present work could easily be adapted to a plethora of contexts, reducing the workload for human experts who can act as supervisors of the job that a pool of synthetic agents can carry out. This pushes forward the boundaries of the contribution that a large language model can provide in an office context and also represents a notable starting point for research in the field of multi-role architecture for everyday tasks.

\bibliographystyle{splncs04}
\bibliography{biblio}

\end{document}

%% file: table_field_1.tex
%\vspace{-5mm}
\begin{table}[H]
\begin{tabularx}{\textwidth}{p{0.7\textwidth}|p{0.3\textwidth}}
\hline
\multicolumn{2}{c}{\textbf{Prompt engineering for the Semantics Identification Agent}} \\ \hline
\multicolumn{2}{l}{\textbf{Template text}: \emph{Your Name}}\\ \hline
\textbf{Agent task} & \textbf{Agent answer} \\ \hline
You are an assistant helping to understand the structure content of a template document. Respond with the content that might be written in the document. & [Your Name]\newline [Your Address]\newline [City, State, Zip Code]\newline [Email Address]\newline [Phone Number] \\ \hline
You are an assistant helping to understand the structure content of a template document. Respond with the content that might be written in the document. Give just the action to do. & Fill in your full legal name.  \\ \hline
You are an assistant helping to understand the structure content of a template document. Respond with the content that might be written in the document. Give just the action to do. For example, if you read "Location", the output will be "Add the location"; if you read "Dear Someone", the output will be "Add salutation". & Add your full name
\\
\hline
\end{tabularx}
\newline
\caption{Agent task prompt refinement for Semantics Identification Agent, with template text "\emph{Your name}"\\.}
\label{tab:semantics-identification-table-1}
\end{table}
%\vspace{-10mm}

%% file: table_field_2.tex
%\vspace{-5mm}
\begin{table}[H]
\begin{tabularx}{\textwidth}{p{0.55\textwidth}|p{0.45\textwidth}}
\hline
\multicolumn{2}{c}{\textbf{Prompt engineering for the Semantics Identification Agent}} \\ \hline
\multicolumn{2}{l}{\textbf{Template text}: \emph{Dear Mr./Ms. (Lastname):}} \\ \hline
\textbf{Agent task} & \textbf{Agent answer} \\ \hline
You are an assistant helping to understand the structure content of a template document. Respond with the content that might be written in the document. &
I hope this message finds you well. I am writing to inform you about the upcoming meeting scheduled for (date) at (time) in (location). The purpose of this meeting is to discuss (agenda items). Your presence and input at this meeting are crucial, so please make every effort to attend. If you are unable to attend, please notify me as soon as possible. Thank you for your attention to this matter, and I look forward to seeing you at the meeting. \\ \hline
You are an assistant helping to understand the structure content of a template document. Respond with the content that might be written in the document. Give just the action to do. &
Insert the recipient's last name in the salutation. \\ \hline
You are an assistant helping to understand the structure content of a template document. Respond with the content that might be written in the document. Give just the action to do. For example, if you read "Location", the output will be "Add the location"; if you read "Dear Someone", the output will be "Add salutation". &
Add salutation\\
\hline
\end{tabularx}
\newline
\caption{Agent task prompt refinement for Semantics Identification Agent, with template text "\emph{Dear Mr./Ms. (Lastname):}".\\}
\label{tab:semantics-identification-table-2}
\end{table}
%\vspace{-15mm}

%% file: table_field_3.tex
%\vspace{5mm}
\begin{table}[H]
\begin{tabularx}{\textwidth}{p{0.45\textwidth}|p{0.55\textwidth}}
\hline
\multicolumn{2}{c}{\textbf{Prompt engineering for the Semantics Identification Agent}} \\ \hline
\multicolumn{2}{p{\linewidth}}{
        \textbf{Template text:} \emph{First paragraphs : In your opening paragraph, concisely convey your enthusiasm for the role
        and the organization, while emphasizing your pertinent educational background. Aim to engage the
        reader's interest, prompting them to delve deeper into your resume. This section serves as a preview,
        showcasing your qualifications and setting the stage for the detailed skills and experiences outlined in
        your resume. Highlight your top skills that align with the job requirements, and showcase your
        familiarity with the organization by mentioning specific aspects such as their mission, ongoing
        projects, or strategic objectives, linking them to how your abilities and interests can add value
        to their team.}
    } \\ \hline
\textbf{Agent task} & \textbf{Agent answer} \\ \hline
You are an assistant helping to understand the structure content of a template document. Respond with the content that might be written in the document. &Dear [Hiring Manager's Name],
I am writing to express my genuine enthusiasm for the [Job Title] position at [Organization Name]. With a strong educational background in [relevant field], I am confident in my ability to contribute to the success of your team. The innovative projects and strategic objectives of [Organization Name] resonate with my professional interests, and I am eager to bring my skills and expertise to support your mission.
My top skills in [specific skills] and [specific skills] align closely
 \\ \hline
You are an assistant helping to understand the structure content of a template document. Respond with the content that might be written in the document. Give just the action to do. &
Express enthusiasm for the role and the organization, emphasizing your educational background and top skills that align with the job requirements. Showcase your familiarity with the organization by mentioning specific aspects such as their mission, ongoing projects, or strategic objectives, and link them to how your abilities and interests can add value to their team.
\\ \hline
You are an assistant helping to understand the structure content of a template document. Respond with the content that might be written in the document. Give just the action to do. For example, if you read "Location", the output will be "Add the location"; if you read "Dear Someone", the output will be "Add salutation". &
Write an enthusiastic opening paragraph that emphasizes your relevant educational background and showcases your qualifications and skills that align with the job requirements. Mention specific aspects of the organization such as their mission, ongoing projects, or strategic objectives, and link them to how your abilities and interests can add value to their team.\\
\hline
\end{tabularx}
\newline
\caption{Agent task prompt refinement for Semantics Identification Agent, with a long and very discursive template text.\\}
\label{tab:semantics-identification-table-3}
\end{table}

%% file: table_missing_1.tex
%\vspace{-5mm}
\begin{table}[H]
\vspace{4mm}

\begin{tabularx}{\textwidth}{p{0.7\textwidth}|p{0.3\textwidth}}
\hline
\multicolumn{2}{c}{\textbf{Prompt Engineering for the Information retrieval Agent}} \\ \hline
\multicolumn{2}{p{\linewidth}}{
    \textbf{Accumulated prompt}:
    "My name is John Doe, i want write a letter for Random University, i am a student in Computer Science"
} \\ \hline
\multicolumn{2}{p{\linewidth}}{\textbf{Section-specific task description}:\newline
\verb|I want you to satisfy this instruction|\newline
[Answer of the Semantics Identification Agent for section "\emph{Your Name}"]\newline
\verb|What is the missing information?|
} \\ 
\hline

\multicolumn{2}{p{\linewidth}}{}
 \\ 
\hline
\textbf{Agent task} & \textbf{Agent answer} \\ \hline
    You are an assistant that has the purpose of searching which information is missing.The information that you have are\newline\newline <\textbf{Accumulated prompt}>\newline\newline
    If you have the information write the token [ALL\_INFO].\newline\newline
     <\textbf{Section-specific task description}>\newline
    &
    The missing information is the full name of the university you want to write the letter to.
 \\ \hline
    You are an assistant that has the purpose of searching which information is missing.The information that you have are \newline\newline <\textbf{Accumulated prompt}>\newline\newline
    Your output must be like this: the missing information to satisfy the request is INFORMATION\_MISSING. If you have the information write the token [ALL\_INFO].
    Strictly respond with only the information that is missing.\newline\newline
     <\textbf{Section-specific task description}>\newline
    &
    The missing information to satisfy the request is [ALL\_INFO].
\\
\hline
\end{tabularx}
\newline

\caption{Agent task prompt refinement for Information retrieval.\\}

\label{tab:missing_information_1}
\end{table}
\vspace{-10mm}

%% file: table_missing_2.tex
%\vspace{-5mm}
\begin{table}[H]
\vspace{5mm}

\begin{tabularx}{\textwidth}{p{0.7\textwidth}|p{0.3\textwidth}}
\hline
\multicolumn{2}{c}{\textbf{Prompt Engineering for the Information retrieval Agent}} \\ \hline
\multicolumn{2}{p{\linewidth}}{
    \textbf{Accumulated prompt}: "My name is John Doe, I want to write a letter for Random University, I am a student in Computer Science"
} \\ \hline

\multicolumn{2}{p{\linewidth}}{\textbf{Section-specific task description}:\newline
\verb|I want you to satisfy this instruction|\newline
[Answer of the Semantics Identification Agent for section "\emph{Dear Mr./Ms. (Lastname):}"]\newline
\verb|What is the missing information?|
} \\
\hline
\multicolumn{2}{p{\linewidth}}{}
 \\ 
\hline
\textbf{Agent task} & \textbf{Agent answer} \\ \hline
You are an assistant that has the purpose of searching which information is missing.The information that you have are\newline\newline <\textbf{Accumulated prompt}>\newline\newline
    If you have the information write the token [ALL\_INFO].\newline\newline
     <\textbf{Section-specific task description}>\newline
    &
    The missing information is the specific name of the person to whom the letter is addressed. Typically, a salutation would include the recipient's name, such as "Dear [Recipient's Name]," or "To Whom It May Concern," if the specific
 \\ \hline
    You are an assistant that has the purpose of searching which information is missing.The information that you have are \newline\newline <\textbf{Accumulated prompt}>\newline\newline
    Your output must be like this: the missing information to satisfy the request is INFORMATION\_MISSING. If you have the information write the token [ALL\_INFO].
    Strictly respond with only the information that is missing.\newline\newline
     <\textbf{Section-specific task description}>\newline    &
    The missing information to satisfy the request are the salutation.
\\
\hline
\end{tabularx}
\newline
\caption{Agent task prompt refinement for Information retrieval.\\ }
\label{tab:missing_information_2}
\end{table}
\vspace{-10mm}

%% file: table_missing_3.tex
\vspace{-5mm}
\begin{table}[H]
\vspace{4mm}
\begin{tabularx}{\textwidth}{p{0.7\textwidth}|p{0.3\textwidth}}
\hline
\multicolumn{2}{c}{\textbf{Prompt Engineering for the Information retrieval Agent}} \\ \hline
\multicolumn{2}{p{\linewidth}}{
    \textbf{Accumulated prompt}: "My name is John Doe, I want to write a letter for Random University, I am a student in Computer Science"
} \\ \hline

\multicolumn{2}{p{\linewidth}}{\textbf{Section-specific task description}:\newline
\verb|I want you to satisfy this instruction|\newline
[Answer of the Semantics Identification Agent for a verbose template body section]\newline
\verb|What is the missing information?|
} \\
\hline
\multicolumn{2}{p{\linewidth}}{}
 \\ 
\hline
\textbf{Agent task} & \textbf{Agent answer} \\ \hline
You are an assistant that has the purpose of searching which information is missing.The information that you have are\newline\newline <\textbf{Accumulated prompt}>\newline\newline
    If you have the information write the token [ALL\_INFO].\newline\newline
     <\textbf{Section-specific task description}>\newline
    &
    The missing information includes your current level of education (e.g., undergraduate, graduate), any specific skills or experiences you want to highlight, and the specific job requirements or aspects of Random University with which you want your skills to align to.
 \\ \hline
    You are an assistant that has the purpose of searching which information is missing.The information that you have are \newline\newline <\textbf{Accumulated prompt}>\newline\newline
    Your output must be like this: the missing information to satisfy the request is INFORMATION\_MISSING. If you have the information write the token [ALL\_INFO].
    Strictly respond with only the information that is missing.\newline\newline
     <\textbf{Section-specific task description}>\newline    &
    The missing information to satisfy the request are the specific aspects of the organization such as their mission, ongoing projects, or any other relevant details.
\\ \hline
\end{tabularx}
\newline
\caption{Agent task prompt refinement for Information retrieval, with a very long and discursive core section of the template document}
\label{tab:missing_information_3}
\end{table}
\vspace{-10mm}